%% file: main_camera.tex
\documentclass[10pt,twocolumn,letterpaper]{article}

\usepackage[pagenumbers]{cvpr} %

\usepackage{graphicx}
\usepackage{amsmath}
\usepackage{amssymb}
\usepackage{booktabs}

\usepackage[pagebackref,breaklinks,colorlinks]{hyperref}

\usepackage[capitalize]{cleveref}
\crefname{section}{Sec.}{Secs.}
\Crefname{section}{Section}{Sections}
\Crefname{table}{Table}{Tables}
\crefname{table}{Tab.}{Tabs.}

\def\cvprPaperID{7918} %
\def\confName{CVPR}
\def\confYear{2022}

\input{preamble}

\begin{document}
\input{sec/0_metadata}
\maketitle
\input{sec/0_abstract}
\input{sec/1_introduction}
\input{sec/2_background}

\input{sec/3_fixedfeature}
\input{sec/4_fullnetwork}
\input{sec/5_largemodel}
\input{sec/6_advtransfer}

\input{sec/7_related}
\input{sec/8_limitations}
\input{sec/9_conclusions}

{
    \small
    \bibliographystyle{ieee_fullname}
    \bibliography{zotero}
}

\input{sec/X_supplementary}

\end{document}

%% file: preamble.tex
\usepackage{overpic}
\usepackage{enumitem} %
\usepackage{overpic} %
\usepackage{color}

\definecolor{turquoise}{cmyk}{0.65,0,0.1,0.3}
\definecolor{purple}{rgb}{0.65,0,0.65}
\definecolor{dark_green}{rgb}{0, 0.5, 0}
\definecolor{orange}{rgb}{0.8, 0.6, 0.2}
\definecolor{red}{rgb}{0.8, 0.2, 0.2}
\definecolor{darkred}{rgb}{0.6, 0.1, 0.05}
\definecolor{blueish}{rgb}{0.0, 0.3, .6}
\definecolor{light_gray}{rgb}{0.7, 0.7, .7}
\definecolor{pink}{rgb}{1, 0, 1}
\definecolor{greyblue}{rgb}{0.25, 0.25, 1}

\newcommand{\Fig}[1]{Fig.~\ref{fig:#1}}

\newcommand{\regular}{Regular}
\newcommand{\ant}{ANT}
\newcommand{\deepaug}{DeepAug+}
\newcommand{\swinT}{Swin-T}

\usepackage{blindtext}

\renewcommand{\paragraph}[1]{\vspace{1em}\noindent\textbf{#1}.}

%% file: sec/0_metadata.tex
\title{Does Robustness on ImageNet Transfer to Downstream Tasks?}

\author{Yutaro Yamada \\
Yale University\\
{\tt\small yutaro.yamada@yale.edu}
\and
Mayu Otani\\
CyberAgent, Inc.\\
{\tt\small otani\_mayu@cyberagent.co.jp}

}

%% file: sec/0_abstract.tex
\begin{abstract}
   As clean ImageNet accuracy nears its ceiling, the research community is increasingly more concerned about robust accuracy under distributional shifts.
   While a variety of methods have been proposed to robustify neural networks, these techniques often target models trained on ImageNet classification.
   At the same time, it is a common practice to use ImageNet pretrained backbones for downstream tasks such as object detection,  semantic segmentation, and image classification from different domains. 
   This raises a question:
   Can these robust image classifiers transfer robustness to downstream tasks?
   For object detection and semantic segmentation, we find that a vanilla Swin Transformer, a variant of Vision Transformer tailored for dense prediction tasks, transfers robustness better than Convolutional Neural Networks that are trained to be robust to the corrupted version of ImageNet.
   For CIFAR10 classification, we find that models that are robustified for ImageNet do not retain robustness when fully fine-tuned. 
   These findings suggest that current robustification techniques tend to emphasize ImageNet evaluations. 
   Moreover, network architecture is a strong source of robustness when we consider transfer learning.
\end{abstract}

%% file: sec/1_introduction.tex
\section{Introduction}
\label{sec:intro}

ImageNet \cite{dengImageNetLargescaleHierarchical2009} serves as an important benchmark in the field of computer vision. 
Numerous models and training techniques have emerged out of this benchmark \cite{heDeepResidualLearning2016a, ioffeBatchNormalizationAccelerating2015}.
A newly proposed vision architecture, including recent Vision Transformer \cite{dosovitskiyImageWorth16x162021}, is first tested against ImageNet to demonstrate a good performance before it gains popularity within the community.
\input{fig/teaser}
While accuracy on ImageNet has been considered as a surrogate for measuring progress in machine vision systems, the research community is now aware of the lack of robustness of vision models towards small input perturbations.
\cite{szegedyIntriguingPropertiesNeural2014a} first reported that imperceptible adversarial perturbations can easily fool image classifiers. 
Recent studies show that even simpler, more natural noises such as blur, contrast change, and snow can significantly degrade the performance of models \cite{hendrycksBenchmarkingNeuralNetwork2019a}.
A typical strategy to increase robustness is data augmentation, where a vision model is trained with additional data, which are artificially corrupted during training.
Examples include ANT \cite{rusakSimpleWayMake2020},  AugMix \cite{hendrycksAugMixSimpleData2020}, and DeepAug \cite{hendrycksManyFacesRobustness2021}. 
However, these techniques often focus on improving robust accuracy for ImageNet classification.
In fact, there are now a variety of ImageNet-scale robustness benchmarks, and the community is striving to improve accuracy on these benchmarks \cite{hendrycksNaturalAdversarialExamples2021, barbuObjectNetLargescaleBiascontrolled2019, hendrycksManyFacesRobustness2021}.

Due to the scale of ImageNet, it is a common practice to use ImageNet pretrained weights for downstream tasks such as object detection \cite{huangSpeedAccuracyTradeOffs2017} and image segmentation \cite{chenDeepLabSemanticImage2018, heMaskRCNN2017}.
This practice of using pretrained ImageNet weights for transfer learning raises a fundamental question from a robustness perspective: When we use pretrained weights that are made to be robust to ImageNet benchmarks, do these models necessarily show robustness for downstream tasks as well? (See Figure \ref{fig:teaser} for the problem setting we consider.)

\paragraph{Contributions}

We find that when we freeze the backbone of ImageNet models, robustified Convolutional Neural Networks (CNNs) maintain robustness for object detection and semantic segmentation.
These robustified CNNs continue to demonstrate higher robustness than the regular model even when we fully fine-tune the whole network, which is practically more relevant.
However, perhaps more notably, we observe that Swin Transformer \cite{liuSwinTransformerHierarchical2021}, a variant of Vision Transformer tailored to dense prediction tasks, transfers robustness better than robustified CNNs in this fully-finetuned setting.
Moreover, it seems difficult to transfer corruption robustness from ImageNet to CIFAR10 \cite{krizhevskyLearningMultipleLayers2009}. 
In fact, we find that a non-robustified ImageNet pretrained ResNet performs the best when fine-tuned for CIFAR10.
We hope these findings encourage the community to reconsider how we evaluate the robustness of vision systems, as existing data augmentation techniques for robustifying neural networks might be overfitting to ImageNet benchmarks. %
Furthermore, it is noteworthy that, for robustness transfer, the robustness contribution from Swin Transformer architecture is more significant than the existing robustification methods.

\paragraph{Scope}
While there are various kinds of distributional shifts and robustness that the vision community studies, we focus on common corruption robustness in this paper, because we are interested in robustness transfer from ImageNet classification to downstream tasks such as object detection and segmentation.
See Section \ref{subsec:bench} for more details about why we specifically choose common corruptions as a topic of our study. 

%% file: fig/teaser.tex
\begin{figure}[t]
\begin{center}
\includegraphics[width=\linewidth]{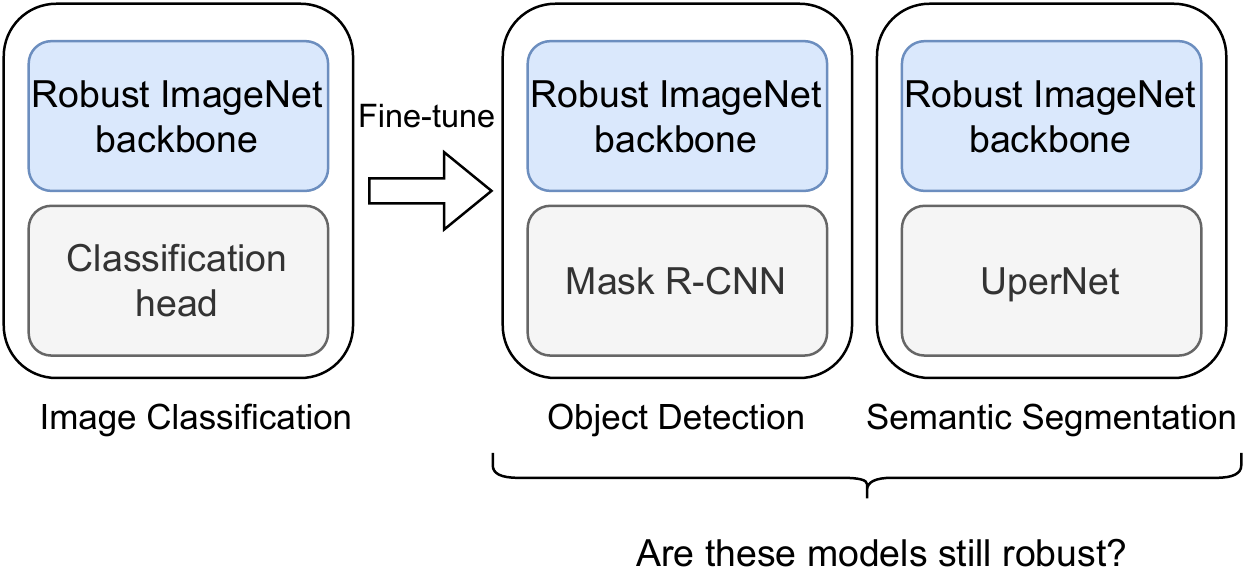}
\end{center}
\caption{
It is a common practice to use ImageNet classifiers as initialization for downstream tasks such as object detection and semantic segmentation.
When we fine-tune robust ImageNet classifiers for downstream tasks, should we expect that the resulting vision system still maintains robustness? 
We tackle this question in the settings of fixed-feature and full-network transfer learning.
} 
\label{fig:teaser}
\vskip -0.2in
\end{figure}

%% file: sec/2_background.tex
\section{Background}

Ensuring robustness in downstream tasks such as object detection and semantic segmentation is equally, if not more, important than achieving robustness in image classification.
Especially for safety-critical applications such as self-driving cars, vision systems that are vulnerable to image perturbations can lead to dire consequences.
In such real-world applications, classification is only the first step of the pipeline, and ensuring robustness through the entire system of object detection and segmentation needs further care.

When we consider how to ensure robustness for downstream tasks, there are two viable approaches.
One is to transfer robustness effectively from a pretrained, robustified classifier backbone to each downstream task, which is our focus of this paper. 
The other approach is to apply an existing robust data augmentation technique during transfer learning.
While applying robustification techniques during finetuning for downstream tasks is an option,
a naive application of these methods can decrease downstream task performances \footnote{See Table \ref{tab:prime_with_and_without_comparison_coco} in the Appendix as an example} and often requires further modifications tailored for downstream tasks to maintain good accuracy while achieving robustness \cite{chenRobustAccurateObject2021}, partly because object detection and semantic segmentation systems tend to be more complex than image classification. 
Therefore, rather than entirely resorting to data augmentation during fine-tuning, it is critical to better understand robustness transfer to achieve both robustness and good clean accuracy in downstream tasks.

\subsection{Vision Transformer for Dense Prediction Tasks}

While image classification only requires a single feature map typically extracted from the last layer, object detection and semantic segmentation benefits a lot from multiresolution feature maps.
These feature maps provide richer information that helps object detection at different scales and pixel-level semantic prediction.
Most object detection and semantic segmentation systems use a CNN as their backbone and exploit hierarchical feature maps that are extracted from different blocks of the model. 

Motivated by the success of Transformer architecture in the Natural Language Processing (NLP) community, Vision Transformer (ViT) \cite{dosovitskiyImageWorth16x162021} was proposed.
While the original ViT excels at image classification, it is not amenable to dense prediction tasks such as object detection and semantic segmentation.
This is because the original ViT processes tokens at fixed scale, producing single low-resolution feature maps.
Recently, a variant of ViT called Swin Transformer was proposed to address this limitation \cite{liuSwinTransformerHierarchical2021}.
Swin Transformer uses a hierarchical architecture to build multiresolution feature maps, while achieving linear-time complexity with respect to the image size.
Because of this, Swin Transformer achieves the state-of-the-art performance in both object detection and semantic segmentation.
In this work, we use Swin Transformer for our ViT architecture.

\subsection{Source of Robustness: Data augmentation and Architecture}

Vanilla CNNs are vulnerable to image corruptions, as extensively studied by the vision community in the past. 
\cite{hendrycksBenchmarkingNeuralNetwork2019a} shows that state-of-the-art ImageNet classifiers fail when naturally occurring image corruptions are applied to the ImageNet test set, which are introduced as ImageNet-C.
To tackle this problem, the community develops many approaches relying on data augmentation \cite{hendrycksManyFacesRobustness2021, xiaoNOISESIGNALROLE2021}.
On the other hand, recent studies show that ViT is more robust to ImageNet-C than vanilla CNNs \cite{naseerIntriguingPropertiesVision2021, bhojanapalliUnderstandingRobustnessTransformers2021} without resorting to data augmentation.
These findings suggest that robustness arises both from data augmentation techniques and architecture itself.
In terms of robustness transfer, it is unclear which source of robustness is more important. 
The following sections explore this question in depth.

%% file: sec/3_fixedfeature.tex
\input{tab/fixedfeature_coco_detect}
\input{tab/fixedfeature_ade_segm}

\section{Fixed-Feature Transfer Learning}
\label{sec:fixedfeature}

When we consider transfer learning from image classifiers to object detection or segmentation, we can freeze the backbone, while only training the head of the detection or segmentation system.
We refer to this approach as fixed-feature transfer learning.
On the other hand, we can use pretrained image classifiers as initialization to train object detection or segmentation models, which we call full-network transfer learning.

Fixed-feature transfer learning from ImageNet to object detection and semantic segmentation is not a common practice because full-network transfer learning generally performs better \cite{huangSpeedAccuracyTradeOffs2017}.
However, for \textit{robustness transfer}, fixed-feature transfer learning is an important setup to consider because it allows us to directly leverage robustified ImageNet backbones and measure how much robustness the model carries over to downstream tasks after fine-tuning only the head of the entire model. 
Full-network transfer learning, on the other hand, potentially erases the robustness property of backbones during fine-tuning, which can confound our analysis of robustness transfer.
 
From earlier work \cite{naseerIntriguingPropertiesVision2021} on robustness of image classifiers, a vanilla Swin Transformer is known to perform better than CNNs on ImageNet-C. 
At the same time, we can robustify these CNNs by data augmentation, so that they perform well on ImageNet-C.
Should we expect that robustified CNNs transfer their robustness automatically when we only fine-tune the head while fixing the backbone?
Moreover, which source of robustness (architecture vs. data augmentation) is better suited in terms of robustness transfer?

To resolve this question, we prepare two CNNs that are robustified during ImageNet-1k pretraining using ANT \cite{rusakSimpleWayMake2020} and DeepAug+AugMix \cite{hendrycksManyFacesRobustness2021} respectively, and a Swin Transformer, also pretrained on ImageNet-1k but without applying any robustification technique.
To control for the model size, we use ResNet50 and Swin-T, where the parameter counts are 25M and 28M, respectively.
For object detection, we use Mask-RCNN \cite{heMaskRCNN2017} and for semantic segmentation, we use UperNet \cite{xiaoUnifiedPerceptualParsing2018} as the head.

\subsection{Robustness Transfer Benchmark}
\label{subsec:bench}

To measure how well a model transfers robustness from ImageNet classification to downstream tasks, we have to prepare the same set of distributional shifts that can be applied to both classification and downstream tasks.
While there are a variety of ImageNet-related benchmarks to measure robustness against distributional shifts (e.g. adversarial \cite{hendrycksNaturalAdversarialExamples2021}, viewpoint change \cite{barbuObjectNetLargescaleBiascontrolled2019}, and background shift \cite{xiaoNOISESIGNALROLE2021}), most of these distributional shifts are not adoptable to our setting, because they are specifically designed for ImageNet classification. 
To measure the performance of robustness transfer to downstream tasks, we focus on 15 synthetic image corruption types, grouped into 4 categories: ``noise'', ``blur'', ``weather'', and ``digital'', introduced in ImageNet-C \cite{hendrycksBenchmarkingNeuralNetwork2019a}.
They measure corruption robustness of ImageNet classifiers by computing how much the original accuracy drops when these models are evaluated on corrupted images of the ImageNet test set. 
Since these image corruptions are algorithmically generated, they can be applied to images in both classification and downstream tasks such as object detection and segmentation. \footnote{We use the following python library to generate synthetic image corruptions \url{https://github.com/bethgelab/imagecorruptions} introduced by \cite{michaelisBenchmarkingRobustnessObject2019}.}
Therefore, these image corruptions allow us to compare the accuracy drop in classification with accuracy drop in downstream tasks, which is useful to measure the degree of robustness transfer across different models.

Formally, we take ImageNet models and fine-tune the head of these models for downstream tasks.
We calculate model performance on the clean test set in downstream tasks, and compute the performance drop after we apply image corruptions.
We then compare the accuracy drop for classification and downstream tasks.
We report the mean performance drop across the 15 image corruptions as our metric. 
The benchmark performance is computed in terms of mean performance under corruption:
\begin{align}
    mPC = \frac{1}{N_c} \sum_{c=1}^{N_c} P_c,
\end{align}
where $N_c$ is 15, and $P_c$ is the task-specific performance measure evaluated under corruption $c$ on the test set.
We then compute the relative performance under corruption:
\begin{align}
    rPC = \frac{mPC}{P_{clean}}
\end{align}
where $P_{clean}$ is the task-specific performance measure evaluated on the clean test set.
We use $1- rPC$ as one of our main metrics to report and refer to this metric as Accuracy Drop or Performance Drop depending on the context. 
$rPC$ allows us to compare the degree of robustness transfer from ImageNet to downstream tasks such as object detection and semantic segmentation.

\paragraph{Dataset}
For object detection, we choose MS-COCO \cite{linMicrosoftCOCOCommon2014} and use the COCO 2017 validation set for COCO as our test split, following the convention.
For semantic segmentation, we choose ADE20K \cite{zhouSceneParsingADE20K2017}. 
ADE20K consists of 20210 train, 2000 validation images, and 150 semantic classes.
For downstream-task specific performance measures, we use the following metrics:

\paragraph{Object Detection}
We use the COCO Average Precision metric, which averages over Intersection-over-Unions (IoUs) between 50\% and 95\%.

\paragraph{Semantic Segmentation}
We use the mean IoU, which indicates the intersection-over-union between the predicted and ground truth pixels, averaged over all the classes.

Table \ref{tab:fixedfeature_coco_detect} and \ref{tab:fixedfeature_ade_segm} summarize the results for the fixed feature transfer learning experiment. 
While ANT and DeepAug+ transfer robustness well across both downstream tasks, we also notice that for some noise types, Swin-T outperforms the robust CNNs (e.g. Noise, Weather in Table \ref{tab:fixedfeature_ade_segm} and Weather in Table \ref{tab:fixedfeature_coco_detect}.)
This suggests that, to our surprise, a vanilla Swin Transformer has a potential to transfer robustness better than robust CNNs.
In the next section, we investigate to what extent these phenomena can be observed in the full-network transfer learning setting.

%% file: tab/fixedfeature_coco_detect.tex
\begin{table}
\centering
\resizebox{0.85\linewidth}{!}{ %
\begin{tabular}{@{}lcccc@{}}
\toprule
Method & Noise & Blur & Digital & Weather \\
\midrule
\regular &  36.09 & 44.00 & 21.17 & 17.59 \\
\ant &  21.90 & 39.25 & 17.70 & 16.22 \\
\deepaug & \textbf{16.39} & \textbf{29.25} & \textbf{15.49} & 11.27 \\
\swinT & 18.01 & 38.18 & 17.90 & \textbf{10.12} \\
\bottomrule
\end{tabular}
} %
\caption{
Performance drops across models and noise types are presented for fixed-feature transfer learning from ImageNet to COCO Object Detection. Regular represents a regular ImageNet-pretrained ResNet50, while DeepAug+ and ANT are ResNet50s that are robustified during ImageNet pretraining. Swin-T is a Swin Transformer (Tiny), where the model size is similar to ResNet50.
} %
\vskip -0.1in
\label{tab:fixedfeature_coco_detect}
\end{table}

%% file: tab/fixedfeature_ade_segm.tex
\begin{table}
\centering
\resizebox{0.85\linewidth}{!}{ %
\begin{tabular}{@{}lcccc@{}}
\toprule
Method & Noise & Blur & Digital & Weather \\
\midrule
\regular &  48.98 & 29.42 & 17.01 & 25.68 \\
\ant &  17.78 & 23.41 & \textbf{12.99} & 25.62 \\
\deepaug & 20.07 & \textbf{19.47} & 13.01 & 19.12 \\
\swinT & \textbf{13.57} & 23.50 & 16.42 & \textbf{14.28} \\
\bottomrule
\end{tabular}
} %
\caption{
Performance drops across models and noise types are presented for fixed-feature transfer learning from ImageNet to ADE10K Semantic Segmentation.
} %
\vskip -0.2in
\label{tab:fixedfeature_ade_segm}
\end{table}

%% file: sec/4_fullnetwork.tex
\input{fig/fullnetwork_comparison}

\section{Full-Network Transfer Learning}
\label{sec:fullnetwork}

A more common practice to perform transfer learning is to use ImageNet pretrained weights as initialization and fine-tune the entire network for downstream tasks.
Even though it takes more computational resources than the fixed-feature case, full-network transfer learning generally performs better \cite{huangSpeedAccuracyTradeOffs2017}.

However, when we take robustness into consideration, full-network transfer learning can be detrimental, because gradient updates during fine-tuning can erase robustified features acquired during ImageNet pretraining. 
This possibility is especially concerning for robustification techniques that rely on data augmentation during pretraining such as DeepAug, AugMix, and ANT.
Thus, one may argue that robustness arising from these data augmentation techniques might be less effective when we fine-tune the entire network for downstream tasks. 
On the other hand, robustness arising from the architecture itself can be more resistant to full-network fine-tuning, because the robustness property is not directly encoded into weights, but rather stems from the topology of architecture.
Thus, we do not need to worry about erasing robustness that arises from architecture during transfer learning. 
As we see that a vanilla Swin Transformer outperforms robustified CNNs for some noise types in the Section \ref{sec:fixedfeature}, architecture indeed plays some role in transferring robustness. 
Therefore, we hypothesize that in the setting of full-network transfer learning, Transformer architectures might be more effective than CNNs that are robustified via data augmentation.

To test this hypothesis, we repeat the same set of experiments as in the Section \ref{sec:fixedfeature}, but now train all weights for object detection, semantic segmentation, and image classification.
For downstream image classification tasks, we choose CIFAR10.
The results are shown in Figure \ref{fig:fullnet_compari}.
As a reference, we also plot the original ImageNet accuracy as well as the Top-1 Accuracy Drop on ImageNet-C for all ImageNet models we use.
We can confirm that the two robust CNNs (DeepAug+ and ANT) indeed demonstrate higher robustness than Regular.
It is noteworthy that a vanilla Swin-T shows slightly higher robustness than ANT (represented as a lower accuracy drop in the blue bar).
More surprisingly, Swin-T performs best in object detection and semantic segmentation.
This shows that DeepAug+ and ANT are less successful to transfer their ImageNet-C robustness to downstream tasks than Swin-T, supporting our hypothesis.
Moreover, when we test robust transfer from ImageNet-C to CIFAR10, we find that these robust models fail to outperform Regular.
This shows that robustness from ImageNet for downstream image classification seems to be harder to transfer than object detection and semantic segmentation.

%% file: fig/fullnetwork_comparison.tex
\begin{figure*}
\begin{center}
\includegraphics[width=\columnwidth]{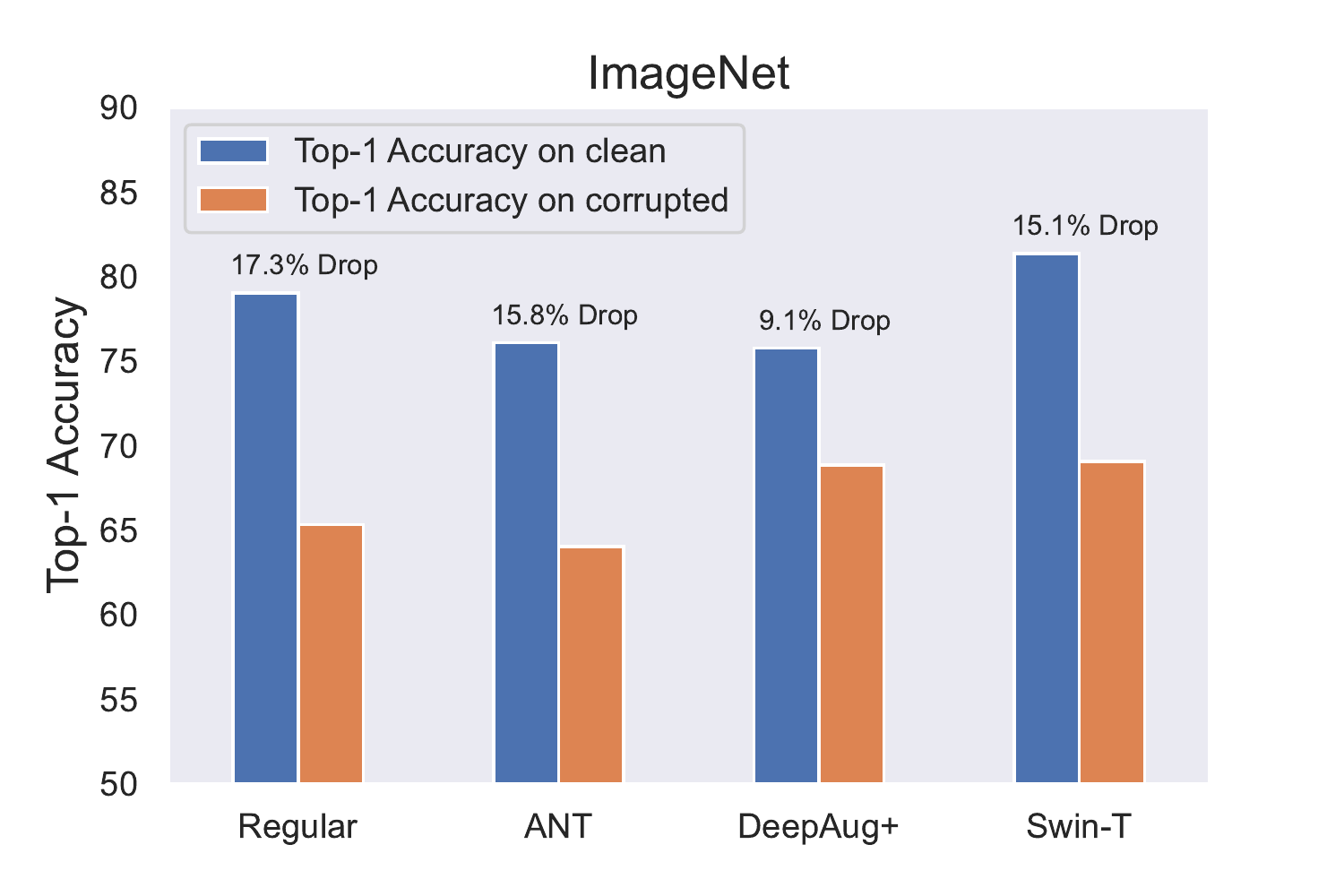}
\hfill
\includegraphics[width=\columnwidth]{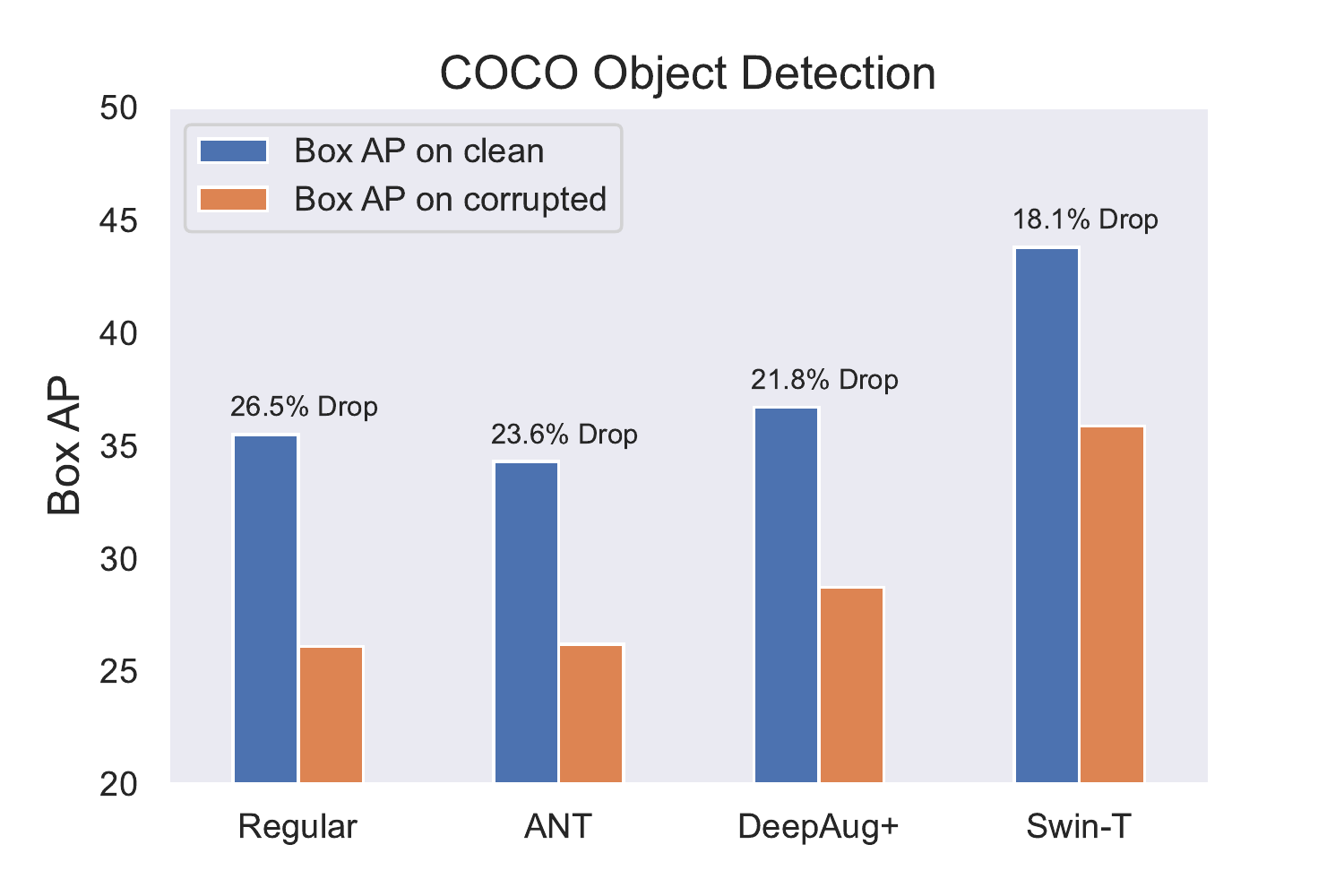}
\hfill
\includegraphics[width=\columnwidth]{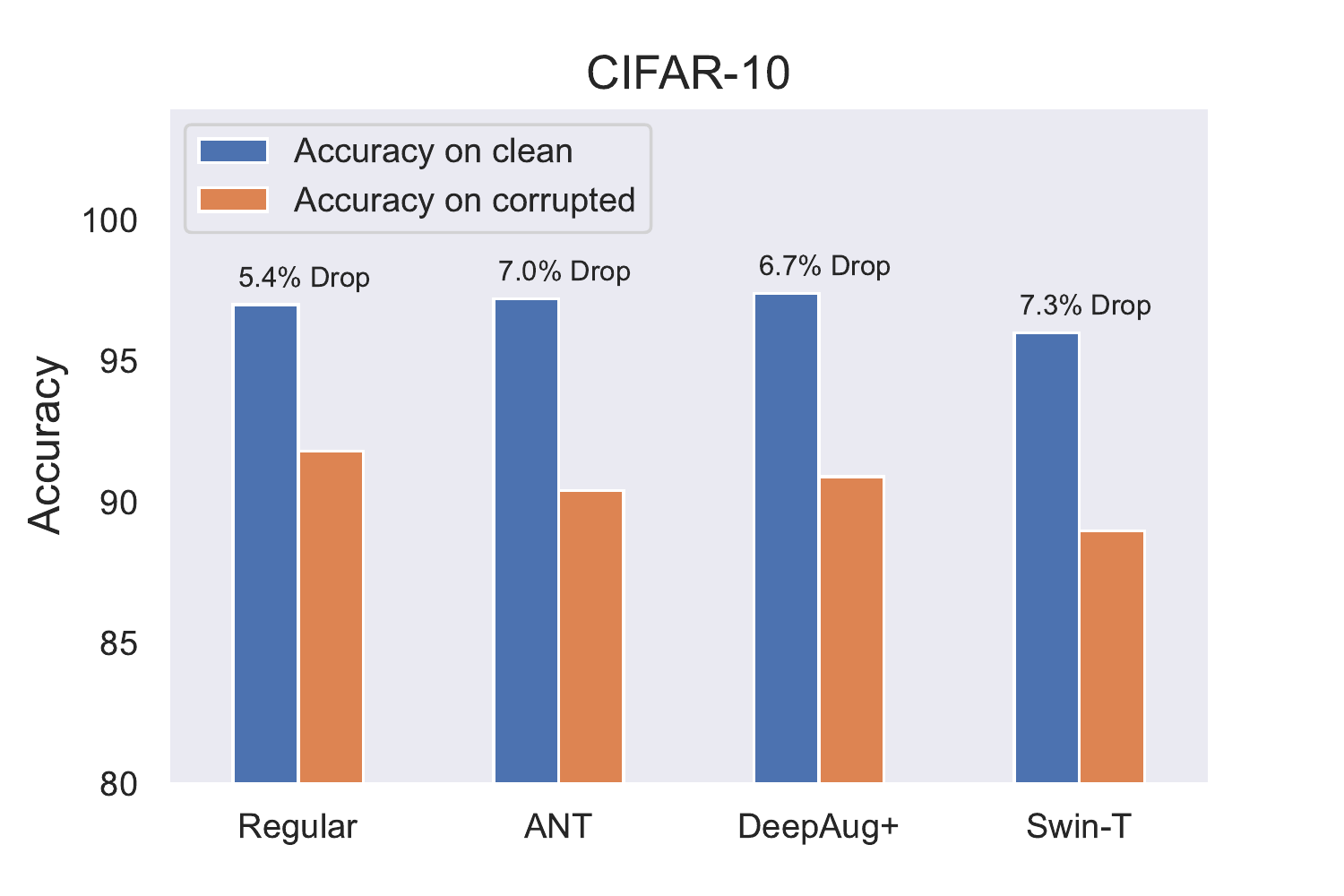}
\hfill
\includegraphics[width=\columnwidth]{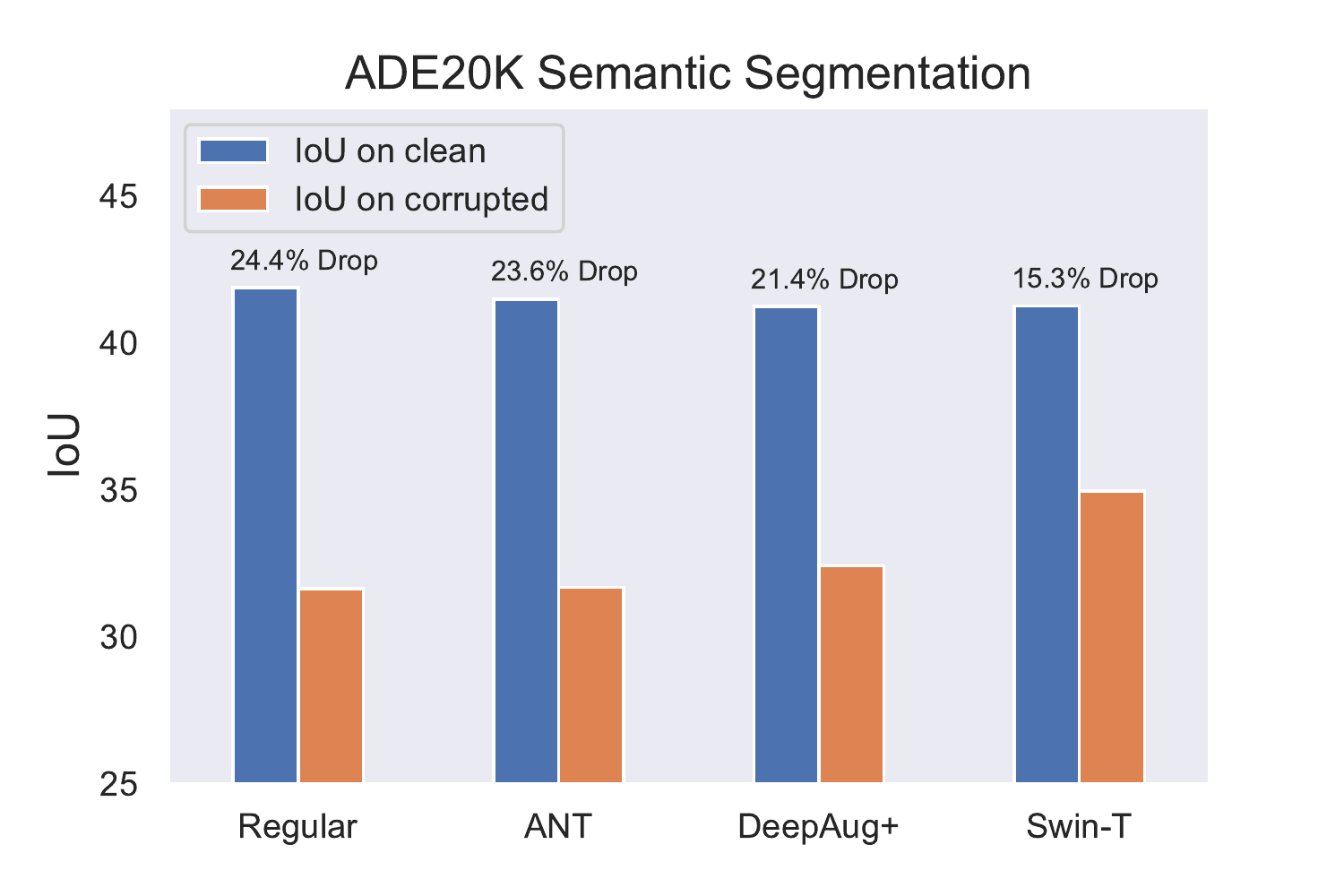}

\end{center}
\caption{
Robust models vs.\ mean performance drop under 15 corruption types in full-network fine-tuning. 
The lower the performance drop, the more robust models are to these image corruptions.
Regular is a vanilla ResNet50.
DeepAug+ and ANT refer to ResNet50 models robustified via DeepAug+AugMix and ANT, which are all data augmentation techniques to increase robustness against common corruptions \cite{hendrycksManyFacesRobustness2021, rusakSimpleWayMake2020}.
Swin-T is a vanilla Tiny Swin Transformer, where the parameter counts are similar to ResNet50.
If robustness on ImageNet is transferable to other downstream tasks, we would see a similar pattern of ImageNet-C in object detection and semantic segmentation as well. 
However, we see that Swin-T performs much better than DeepAug+, the most robust model against ImageNet-C.
This shows that the Swin Transformer as architecture is a stronger source of robustness transfer than robustification techniques that are used (e.g. DeepAug, AugMix, or ANT).
Moreover, for CIFAR-10, Regular appears to be the most robust model, highlighting the difficulty of transferring ImageNet robustness effectively.}
\label{fig:fullnet_compari}
\vskip -0.2in
\end{figure*}

%% file: sec/5_largemodel.tex
\input{tab/swin_vary_model}
\input{fig/abe_largemodel}

\section{Do Larger Models Transfer Robustness Better?}

Having established that the Swin Transformer architecture is a strong source of robustness transfer for object detection and semantic segmentation, especially in full-network transfer learning, we now explore whether or not the size of Transformer architecture affects the performance of model robustness for downstream tasks.
In this section, we compare Tiny, Small, and Base Swin Transformers in full-network transfer learning, where the detailed configurations are shown in Table \ref{tab:swin_model_size}.

Figure \ref{fig:abe_largemodel} shows the mean performance drop after we apply image corruptions as well as the original performance of each model for ADE20K semantic segmentation, COCO object detection, and COCO instance segmentation.
We see that in general the larger the model size is, the smaller the performance drop becomes.
This suggests that larger models tend to have more robustness. 
However, we also observe that there are a few exceptions to this general trend.
For instance, Base in ADE20K Semantic Segmentation and Base in COCO Object Detection and Instance Segmentation demonstrate larger performance drop compared to Small. 
Here we note that the original performances of these large models are similar to Small.
We hypothesize that the failure of these large models can be attributed to the pretrained Swin Transformer models, which only use ImageNet-1k for pretraining.  
Indeed, when Base is pretrained on ImageNet-22k instead of ImageNet-1k, we see that the IoU Performance Drop is smaller than Small. 
Similar phenomena are also reported in \cite{bhojanapalliUnderstandingRobustnessTransformers2021}, where large models tend to require more training data to outperform smaller models on clean test sets. 

%% file: tab/swin_vary_model.tex
\begin{table}
\centering
\resizebox{\linewidth}{!}{ %
\begin{tabular}{@{}lcccc@{}}
\toprule
Model & \#params & Pre-train Data & Input size & window \\
\midrule
Tiny  & 29M & IN-1k & 224 & 7 \\
Small & 50M & IN-1k & 224 & 7 \\
Base'' & 88M & IN-22k & 224 & 7 \\
Base' & 88M & IN-22k & 384 & 12 \\
Base & 88M & IN-1k & 224 & 7\\
\bottomrule
\end{tabular}
} %
\caption{
Swin Transformer architectures we use to test common corruption robustness on object detection and semantic segmentation. The pre-training data is either the ImageNet-1K or ImageNet-22k training set.
} %
\label{tab:swin_model_size}
\end{table}

%% file: fig/abe_largemodel.tex
\begin{figure*}
\begin{center}
\includegraphics[width=.99\columnwidth]{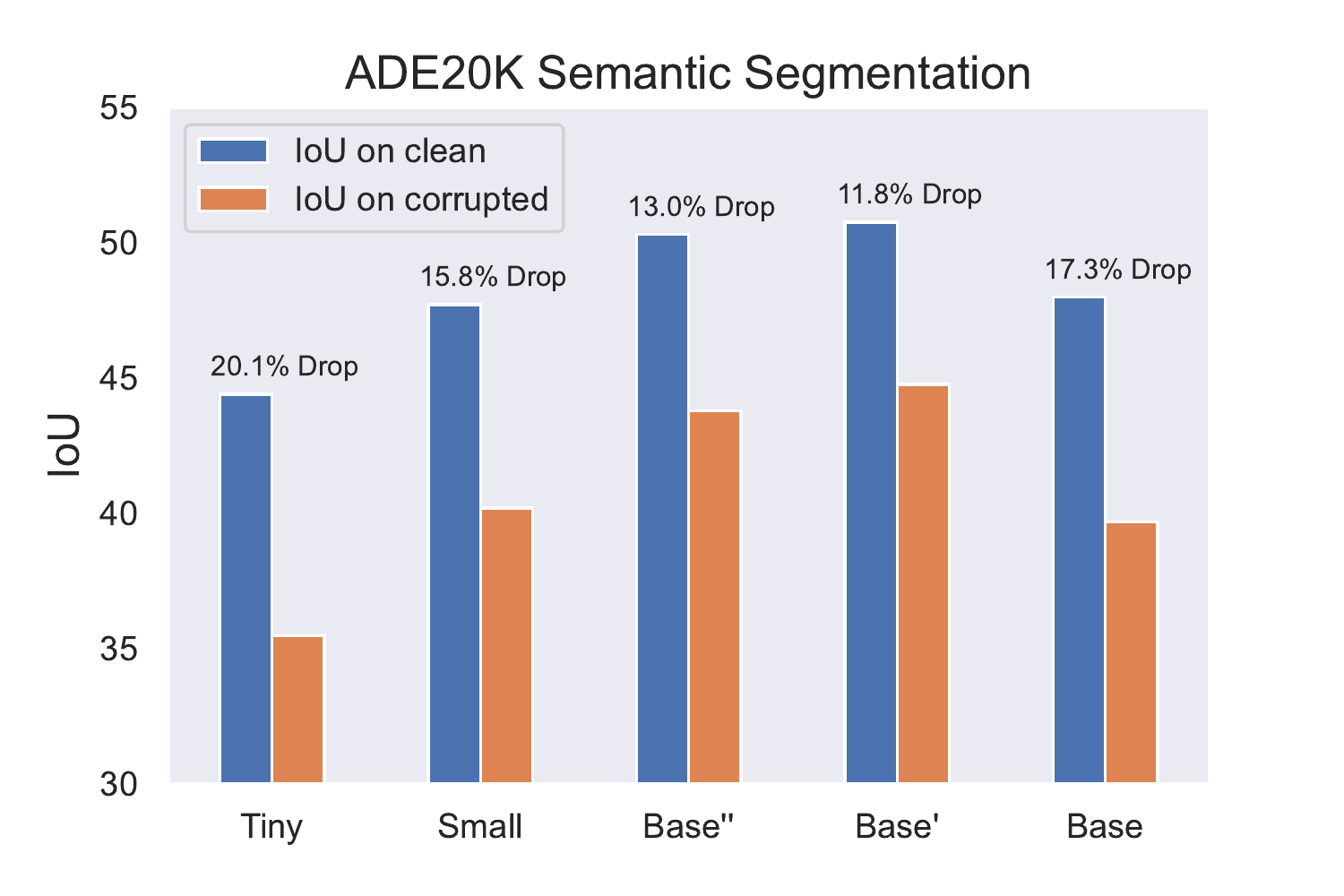}
\hfill
\includegraphics[width=.99\columnwidth]{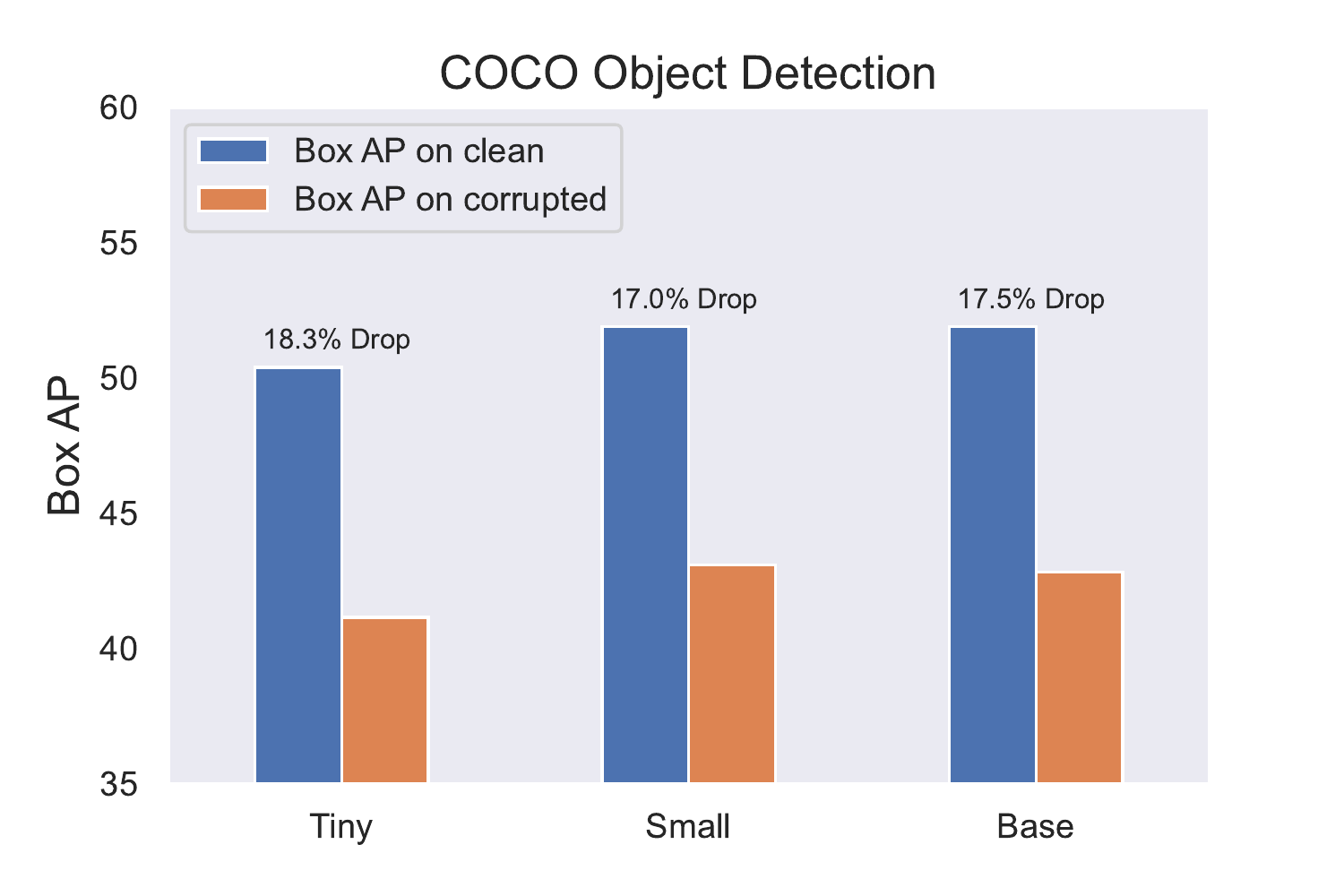}
\hfill
\includegraphics[width=.99\columnwidth]{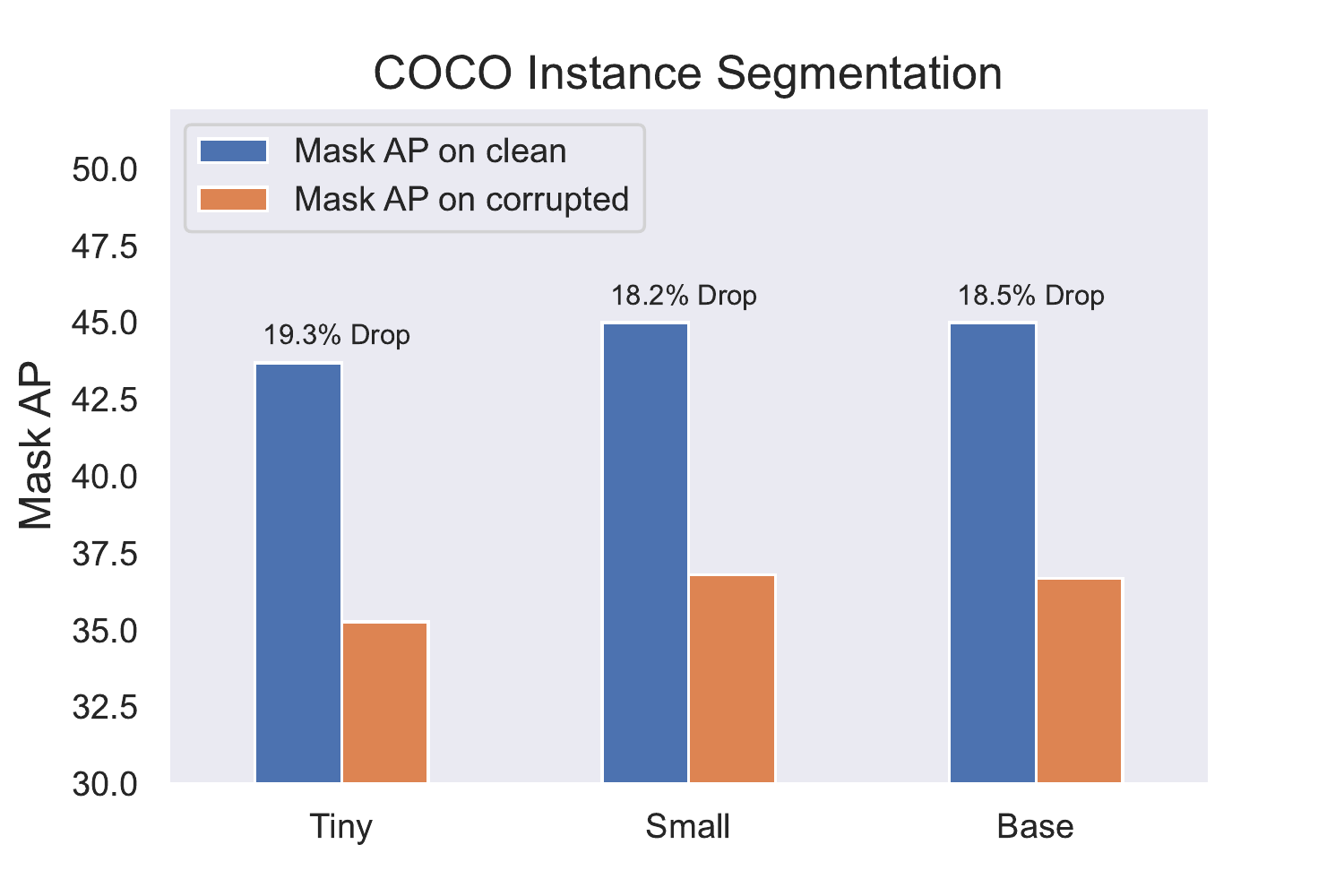}
\end{center}
\caption{
Swin Transformers varying model size vs.\ performance drop for downstream tasks under image corruptions.
Tiny, Small, and Base are all pretrained on ImageNet-1k while Base' and Base'' are pretrained on ImageNet-22k. See Table \ref{tab:swin_model_size} for more details about the difference in configurations of models.
We see that the larger the models, the more robust in general.
However, when we compare Small and Base, it is clear that Base underperforms Small in terms of both robustness to corruption and clean performance. 
This can be attributed to the pretraining dataset size, where Base requires larger training data to regularize the model than Small.
}
\label{fig:abe_largemodel}
\vskip -0.2in
\end{figure*}

%% file: sec/6_advtransfer.tex
\input{fig/abe_adv_eps}

\section{Adversarially-trained Networks do not Transfer Robustness to Downstream Tasks}

Recent studies \cite{salmanAdversariallyRobustImageNet2020, utreraAdversariallyTrainedDeepNets2021} find that adversarial robustness is a good prior for transfer learning.
Adversarial robustness refers to a model's stability against small worst-case input perturbations, called adversarial examples \cite{szegedyIntriguingPropertiesNeural2014a}. 
Robustness is typically induced by training a model on adversarial examples via the following robust optimization objective \cite{madryDeepLearningModels2018}:
\begin{align*}
    \min_{\theta} \mathbb{E}_{x,y \sim \mathcal{D}} \left [ \max_{||\delta||_2 \le \epsilon} \mathcal{L}(x+\delta, y; \theta) \right ],
\end{align*}
where $\theta$ is the model parameter, the expectation is taken over the data distribution $\mathcal{D}$, and $\epsilon$ controls the magnitude of adversarial perturbation vector $\delta$.
Therefore, the larger the $\epsilon$ is, the more robust the adversarially-trained models become.
Their hypothesis was that adversarially-trained networks maintain better-behaved gradients \cite{tsiprasRobustnessMayBe2019, zhangInterpretingAdversariallyTrained2019a}, which might help transfer learning. %

While \cite{salmanAdversariallyRobustImageNet2020, utreraAdversariallyTrainedDeepNets2021} demonstrate that adversarially-trained networks can transfer better than standard models for downstream image classification tasks (without any image corruption), it is unclear how these networks perform in terms of robustness transfer when we consider the performance under image corruptions.
In this section, we investigate if adversarial prior is helpful for robustness transfer from image classification to object detection and semantic segmentation. 
A limitation is that preparing adversarially-trained models from scratch is difficult since adversarial training is resource intensive.
Fortunately,  ResNet50 models that are adversarially-trained on ImageNet are made publicly available by \cite{salmanAdversariallyRobustImageNet2020}.
Here, we focus our study on these ResNet50 pretrained models, and will leave for future work how adversarial prior affects Swin Transformer's robustness transfer.

We use four $\ell_2$-robust models that are trained using $\epsilon = 0.1, 1.0, 3.0, $ and $5.0$, respectively, and fine-tune the whole network for COCO object detection and ADE20K semantic segmentation.
The original ImageNet clean accuracy as well as mean accuracy drop on ImageNet-C are shown in the top-left panel of Figure \ref{fig:abe_adv_eps}.
We see a clear trend that more robust models tend to perform worse on both clean ImageNet and ImageNet-C.
Therefore, the model with $\epsilon=0.1$ is optimal in terms of both clean accuracy and corruption robustness.
For downstream tasks, we see that more robust models perform worse on clean data, but the performance drop is less severe.
In fact, for all downstream tasks we consider, the $\epsilon=0.1$ model performs worst in terms of performance drop on both COCO and ADE20K.
This suggests that adversarial prior of ImageNet classifiers is not helpful for robustness transfer to downstream tasks, as opposed to the regular transfer learning setting, where they evaluate clean performance on downstream image classification tasks.

%% file: fig/abe_adv_eps.tex
\begin{figure*}
\begin{center}
\includegraphics[width=.99\columnwidth]{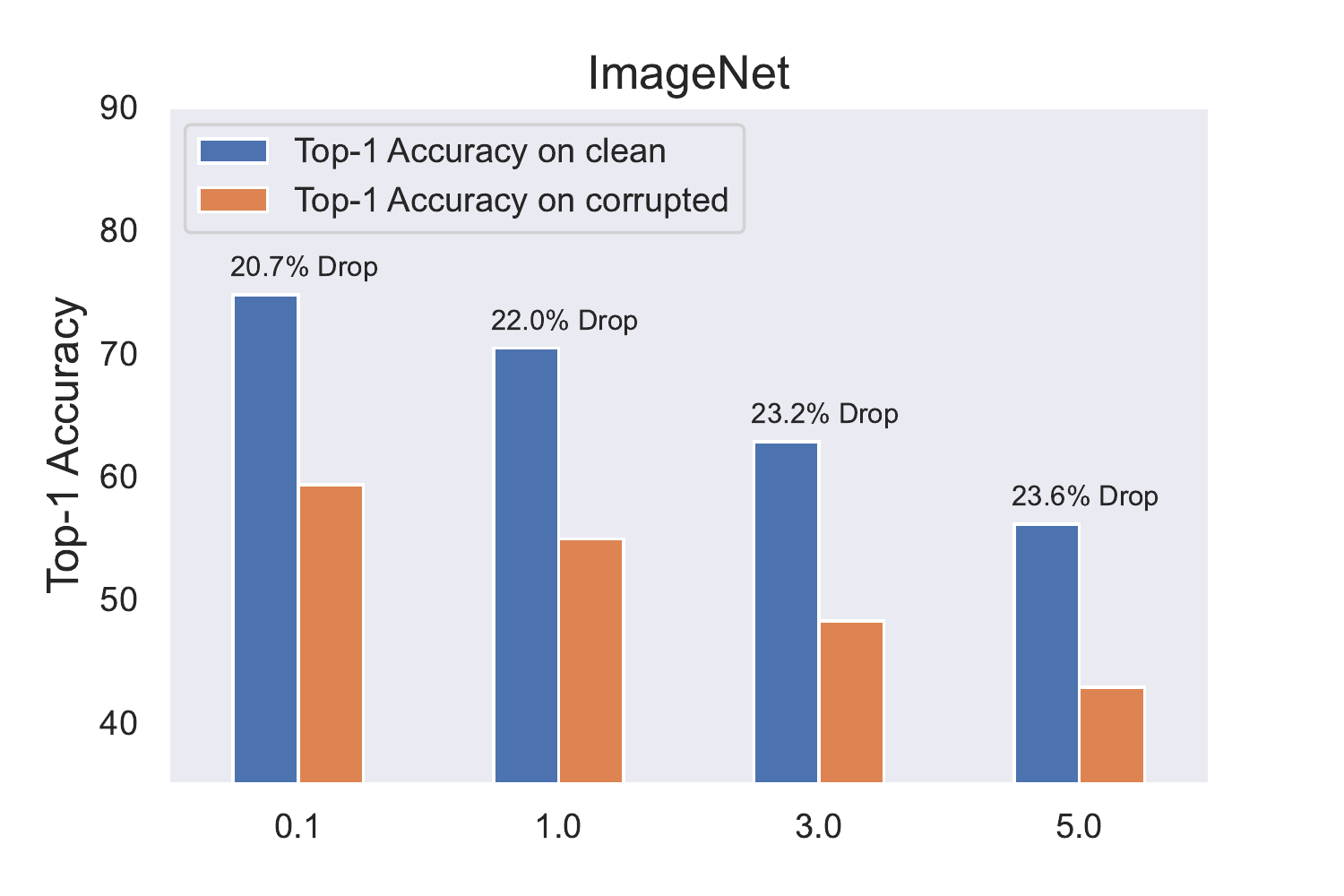}
\hfill
\includegraphics[width=.99\columnwidth]{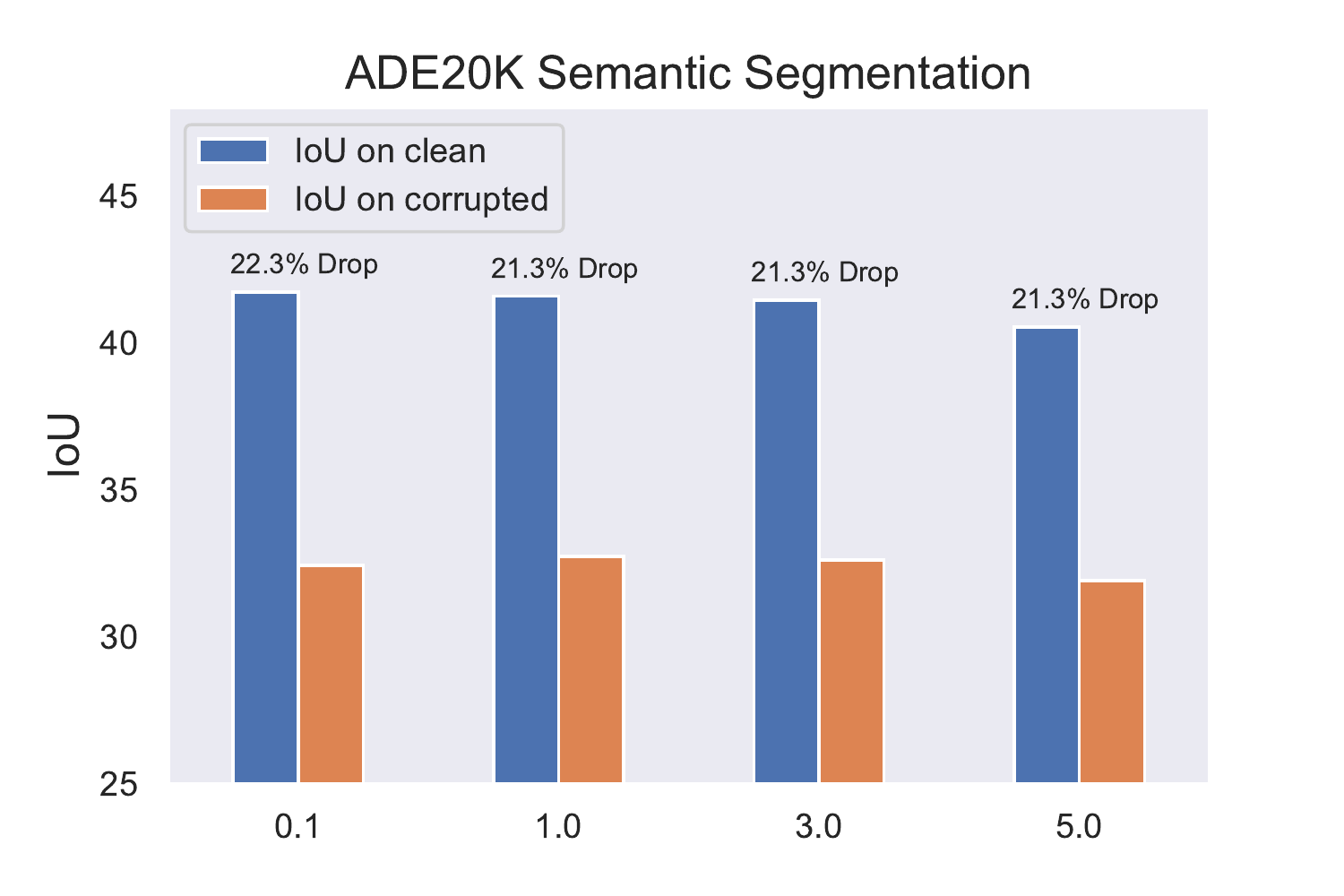}
\hfill
\includegraphics[width=.99\columnwidth]{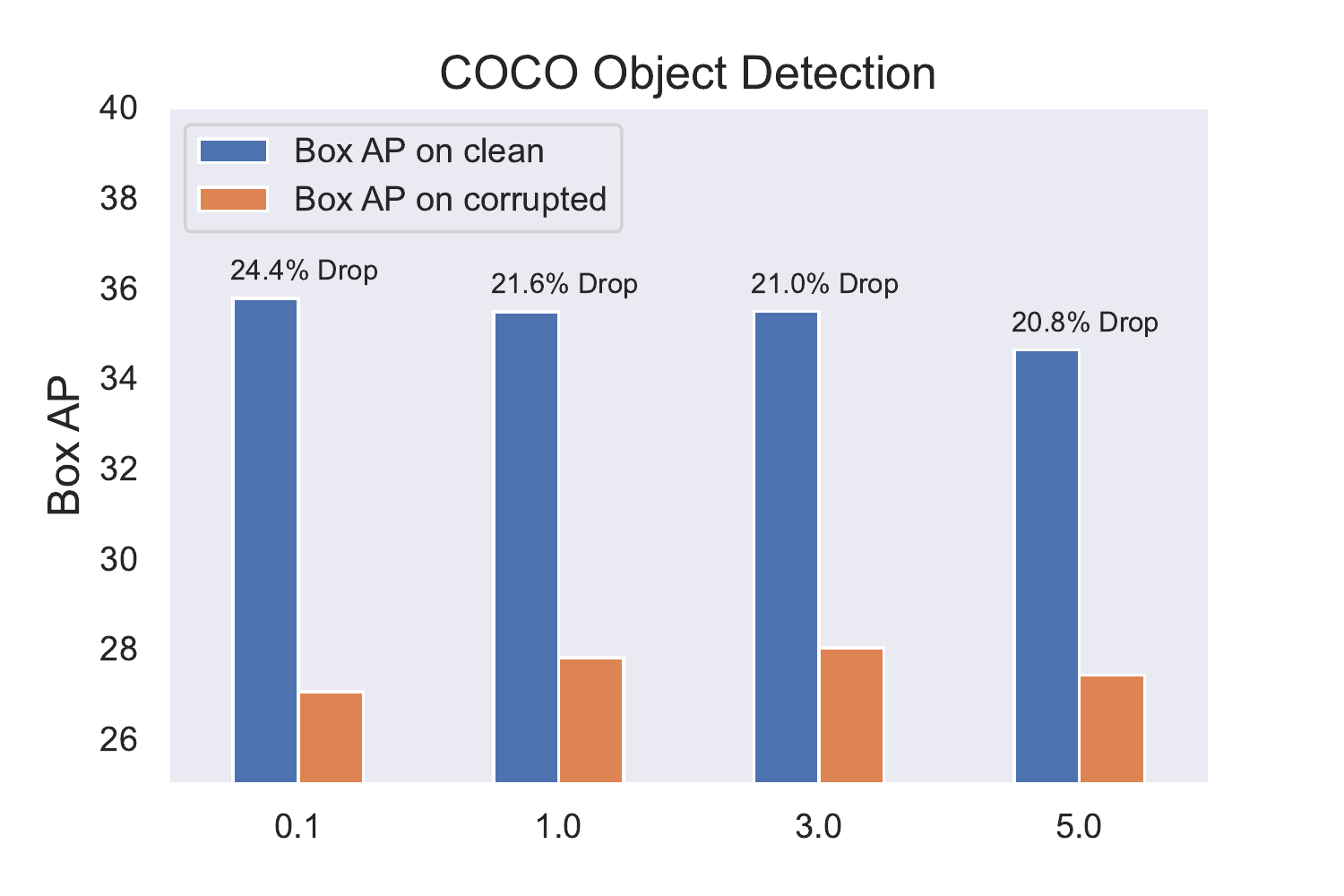}
\hfill
\includegraphics[width=.99\columnwidth]{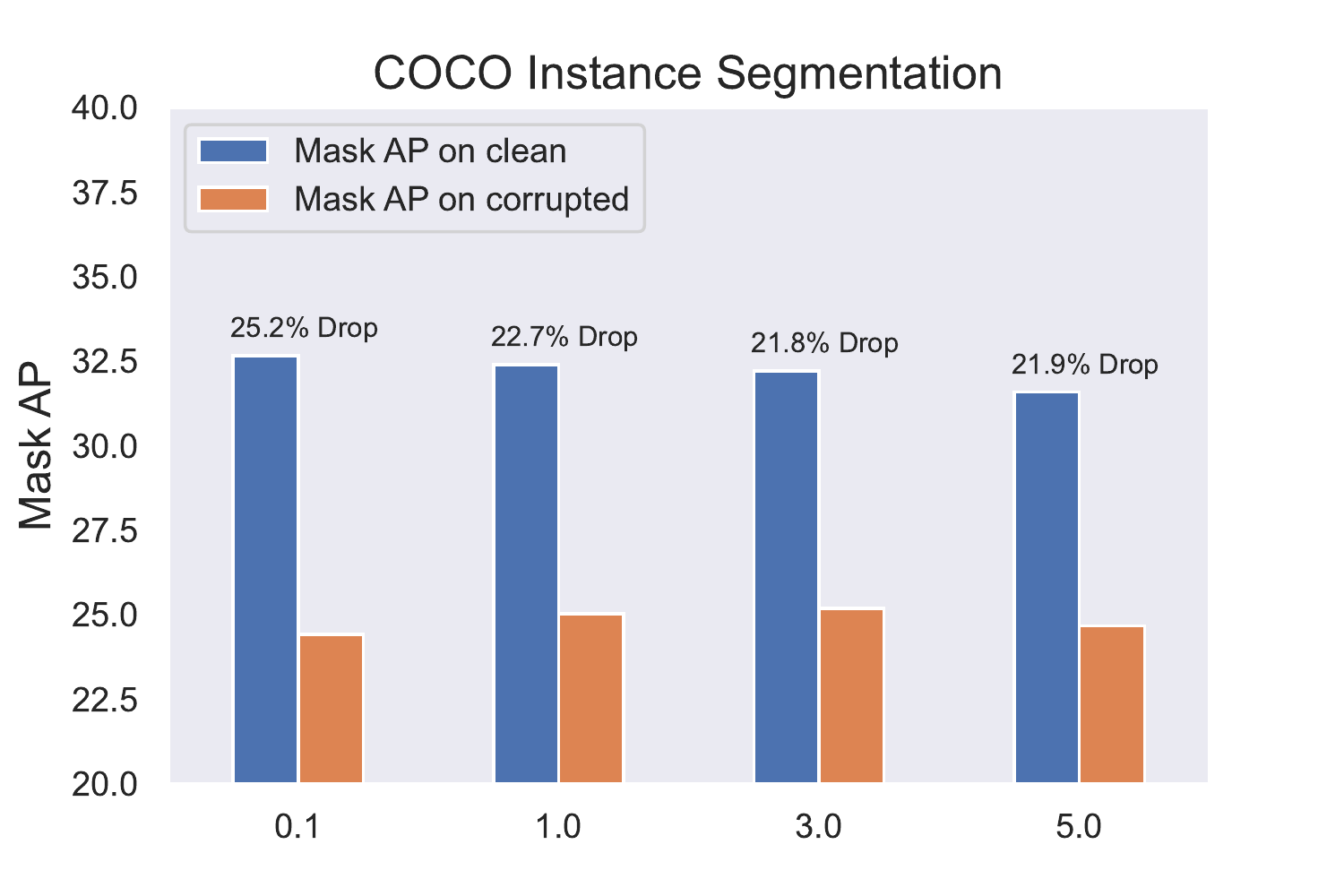}
\end{center}
\caption{
Adversarial robustness of ImageNet classifiers vs.\ performance drop for downstream tasks under image corruptions.
We vary the $\ell_2$-robustness budget $\epsilon$ that is used for adversarial training. The higher the $\epsilon$, the more robust the trained models are towards adversarial attacks in image classification. 
We can see that the model with $\epsilon=0.1$ shows the strongest robustness on ImageNet-C, while the same model reveals the worst robustness when fine-tuned for downstream tasks. 
This shows that adversarial prior is not helpful for robustness transfer as opposed to the findings in \cite{salmanAdversariallyRobustImageNet2020, utreraAdversariallyTrainedDeepNets2021}, where they show adversarial prior is important for transfer learning of image classification.
}
\label{fig:abe_adv_eps}
\vskip -0.2in
\end{figure*}

%% file: sec/7_related.tex
\section{Related Works and Discussions}
\label{sec:related}

\paragraph{Transfer learning to image classification tasks}
\cite{kornblithBetterImageNetModels2019} performs a large-scale study of transfer learning from ImageNet to other image classification tasks.
While they only test CNN architectures, they demonstrate that architectures that perform better on ImageNet are capable of learning better features that are transferable across different classification tasks.
On the other hand, they also show that ImageNet pretrained weights do not necessarily transfer well to small fine-grained classification datasets.
Our findings add to their results in that architecture is not only beneficial for regular transfer learning but also can be a good source of \textit{robustness transfer}.

\cite{huangSpeedAccuracyTradeOffs2017, chenDeepLabSemanticImage2018} demonstrate that ImageNet models with higher accuracy transfer better to object detection and semantic segmentation. 
While these works offer important insights regarding regular transfer learning, our study is orthogonal to these works, because we are specifically interested in how well ImageNet pretrained classifiers can transfer their robustness, instead of clean performance transfer.

\paragraph{Robustness of Vision Transformer}
Our work is inspired by recent findings that Vision Transformers show more robustness than CNNs to common image corruptions \cite{naseerIntriguingPropertiesVision2021}.
\cite{bhojanapalliUnderstandingRobustnessTransformers2021} shows that larger Vision Transformers require more training data to be robust to ImageNet-C.
This is in line with our finding that larger Swin Transformers need ImageNet-21K pretraining to increase robustness in object detection and semantic segmentation.   
It is hypothesized that Vision Transformers generally require more training data than CNNs since they do not have the inductive bias like convolutions that are useful for image domains \cite{dosovitskiyImageWorth16x162021}. 

\paragraph{Robust object detection and segmentation}
There are several studies that attempt to increase adversarial robustness of object detection systems. 
\cite{zhangAdversariallyRobustObject2019} relies on adversarial training.
However, adversarial perturbations are artificially crafted examples. 
A more natural situation is object detection under occlusion.
\cite{wangRobustObjectDetection2020} addresses such an issue by developing specifically designed architectures to handle occlusion.
There are several works that attempt to increase robustness for semantic segmentation. 
\cite{barRobustSemanticSegmentation2020} proposes a specialized student-teacher architecture for robust semantic segmentation.
\cite{kamannIncreasingRobustnessSemantic2020} relies on increasing shape bias of networks to build the robust semantic segmentation system, inspired by the success of image classification using a similar shape-bias approach \cite{geirhosImageNettrainedCNNsAre2019}.
Instead of inducing robustness directly in object detection or semantic segmentation, we study robustness transfer from ImageNet-pretrained models to these downstream tasks.
While developing robust detection or segmentation systems is important, we think it is also beneficial to tackle building robust systems from the point of view of transfer learning from robust image classifiers.

\cite{kamannBenchmarkingRobustnessSemantic2020} benchmarks the robustness of various CNN models for semantic segmentation under image corruptions similar to ImageNet-C.
They find that, within the CNN models they tested, models with higher accuracy show stronger robustness on semantic segmentation.
Our work is orthogonal to their finding as we start from robust models, and how much robust transfer occurs via fine-tuning.

\paragraph{Robustness Transfer}
While there exist prior works on robustness transfer, they focus on transferring adversarial robustness from source models to target models in image classification settings. 
For instance, \cite{shafahiAdversariallyRobustTransfer2020} proposes lifelong learning strategies to transfer adversarial robustness effectively.
\cite{chanWhatItThinks2020} shows that adversarial robustness transfer can be achieved by input gradient adversarial matching in the form of student-teacher framework.
While these works are important, more practically relevant is the task of robustness transfer for common image corruptions such as noise, blur, and weather change.
Furthermore, as opposed to these works, we focus on robustness transfer from image classification to object detection and semantic segmentation.

\paragraph{Overfitting to ImageNet}
Our results from fixed-feature transfer learning suggest that robustness of ImageNet-pretrained backbones can be maintained if we freeze the weights of the backbones, but this sacrifices the validation accuracy on downstream tasks.
A more ideal scenario would be to preserve robustness even in full-network fine-tuning.
However, as Section \ref{sec:fullnetwork} shows, robustness from weights of CNNs were less effective than robustness of the Swin Transfromer architecture.
That is, the robustness performance on ImageNet-C is not perfectly transferable to downstream tasks.
This result is reminiscent of several reports regarding overfitting to ImageNet.
For instance, \cite{rechtImageNetClassifiersGeneralize2019b} studies that ImageNet models do not generalize well to additional test data generated using a data collection process similar to that of ImageNet.
Another work \cite{shankarImageClassifiersGeneralize2021} shows that ImageNet pretrained models do not generalize to videos.
Our robustness transfer results add to these works, suggesting again that over-reliance on ImageNet benchmarks can be misleading. %

%% file: sec/8_limitations.tex
\section{Limitations}

While we claim that the architecture is a strong source of robustness for transfer learning, this statement is limited in a sense that we only compare ResNet and Swin Transformer. 
We encourage future work to study broader types of architectures and what properties of models can be well-preserved during transfer learning.
We also note that our goal is to study the effect of robust transfer, and therefore we did not necessarily aim for achieving the state-of-the art performance on downstream tasks.
Developing a general recipe for achieving good clean accuracy while maintaining robustness on downstream tasks remains an important future work.

%% file: sec/9_conclusions.tex
\section{Conclusions}

In this work, we study the problem of \textit{robustness transfer} from ImageNet pretrained classifiers
to downstream tasks such as object detection and semantic segmentation.
Our study is motivated by the two observations: 
1.\ Even though there are many proposals to robustify neural networks, 
these methods target ImageNet classifiers. 
2.\ It is common to use ImageNet pretrained weights for object detection and semantic segmentation. 
This leads to our central question of this paper: Do robustified ImageNet classifiers necessarily transfer robustness to downstream tasks?
In the fixed-feature transfer learning setting, we find that robustness of ImageNet backbones is partially preserved on downstream tasks.
However, in full-network transfer learning, which is more practically relevant, we see that 
the contribution from the Transformer architecture is more significant than the specific robustification techniques that are applied to CNNs.
We also test if the adversarial prior, which is shown to be important for regular transfer learning, is also important for robustness transfer. 
We find that, as opposed to previous findings, the adversarial prior does not help robustness transfer.
We hope that our findings encourage the community to reconsider how we evaluate corruption robustness of vision systems.

%% file: sec/X_supplementary.tex
\appendix

\setcounter{page}{1}

\twocolumn[
\centering
\Large
\textbf{Does Robustness on ImageNet Transfer to Downstream Tasks?} \\
\vspace{0.5em}Supplementary Material \\
\vspace{1.0em}
] %
\appendix

\section{Training Details}

\paragraph{Object Detection on MS-COCO 2017}
For Swin Transformer, we employ the setting used in \cite{liuSwinTransformerHierarchical2021}.
For ResNet50-based models, we train Mask-RCNN using the default model and hyperparameter configurations from mmdetection \footnote{\url{https://github.com/open-mmlab/mmdetection/blob/master/configs/_base_/models/mask_rcnn_r50_fpn.py}, \url{https://github.com/open-mmlab/mmdetection/blob/master/configs/_base_/datasets/coco_instance.py}},
except for the learning rate of SGD, which we set to 0.04 based on the grid-search over 0.01, 0.04, and 0.08.
We replace random initialization with pretrained weights from DeepAug+AugMix, ANT, or adversarially-trained models.

\paragraph{Semantic Segmentation on ADE20K}
For Swin Transformer, we employ the setting used in \cite{liuSwinTransformerHierarchical2021}.
For ResNet-50-based models, we train UperNet using the default model and hyperparameter configurations from mmsegmentation \footnote{\url{https://github.com/open-mmlab/mmsegmentation/blob/master/configs/upernet/upernet_r50_512x512_80k_ade20k.py}, \url{https://github.com/open-mmlab/mmsegmentation/blob/master/configs/_base_/models/upernet_r50.py}, \url{https://github.com/open-mmlab/mmsegmentation/blob/master/configs/_base_/datasets/ade20k.py}, \url{https://github.com/open-mmlab/mmsegmentation/blob/master/configs/_base_/schedules/schedule_80k.py}},
except for the backbone type, which we set to ResNet50.
We replace random initialization with pretrained weights from DeepAug+AugMix and ANT.

\paragraph{Image Classification on CIFAR10}
We train all models using the same configuration.
Following the recommendations made by \cite{salmanAdversariallyRobustImageNet2020}, we use batch size of 64, momentum of 0.9, weight decay of 5e-4 and the learning rate of 0.01, which we reduce by a factor of 10 every 50 epochs.

\section{Additional Experiments}

\subsection{Degree of fine-tuning}

In this section, we study how the degree of fine-tuning affects the performance of robust transfer for the full-network transfer learning, where we use pretrained, robustified ImageNet weights as our initialization.
It is reasonable to expect that when we use a small learning rate for fine-tuning, we should be able to retain some of the robustness properties that the robustified ImageNet models have.
On the other hand, small learning rates slow down training processes and can fail to reach convergence under limited computational resources.
To investigate this trade-off, we train PRIME-pretrained ImageNet models with varying learning rates under the same computational budget.
The results are shown in \Fig{abe_degree_of_finetune}.
As expected, the smaller the learning rate, the lower the clean performance becomes, and consequently the raw performance on the corrupted data also decreases at least for the COCO object detection task.
However, the percentage of the performance drop becomes less significant for the small learning rate cases.
This partially confirms our intuition.
We leave as future work maintaining robustness while improving performance on downstream tasks for transfer learning.

\input{fig/abe_degree_of_finetune}

\subsection{Data augmentation during fine-tuning}

While we focus on studying how much robustness transfer learning can retrain from robust pretrained ImageNet models to downstream tasks, it is also reasonable to simply use robustification techniques during transfer learning since our ultimate goal is to have a robust model on downstream tasks.
We use PRIME \cite{modasPRIMEFewPrimitives2022} as a data augmentation technique that is intended to robustify image classification models.
We chose this method because it does not require to modify the loss function unlike ANT and AugMix, and is much simpler than DeepAug.
The results are shown in \Fig{abe_data_aug_while_finetune}.
We use three ImageNet pretrained weights to initialize ResNet50, where Regular is a standard ImageNet pretrained weights, and DeepAug+ and PRIME indicate that the corresponding ImageNet training incorporates either DeepAug+ or PRIME data augmentation.
Swin-T is pretrained on ImageNet, but without any robustification technique during pre-training. 
For transfer learning from Swin-T, we use PRIME.

In Table \ref{tab:prime_with_and_without_comparison_coco} and \ref{tab:prime_with_and_without_comparison_ade20k}, we compare the effect of the PRIME data augmentation during transfer learning.
While applying PRIME during transfer learning mitigates the performance drop from the clean data to corrupted data, we can also see that it significantly hurts the clean performance on COCO.
Interestingly, for ADE20K semantic segmentation, applying PRIME during fine-tuning slightly improves the clean performance.

In Figure \ref{fig:abe_data_aug_while_finetune}, we compare the effect of the PRIME data augmentation during transfer learning across models.
Here again we can observe that PRIME hurts the clean performance for COCO but does not significantly affect ADE20K.
Even in the PRIME fine-tuning setting, we can see that Swin-T performs the best among the models we evaluated.

We note that for PRIME during transfer learning, we omit the geometric transformation module from PRIME so that the PRIME data augmentation does not distort geometric information that is tied with object bounding boxes and semantic labels in ADE20K and COCO.
Furthermore, we use the default hyperparameters of PRIME that are designed for ImageNet, and did not incorporate the JSD consistency loss in the PRIME module.
More careful tuning of PRIME and incorporation of the JSD consistency loss into object detection / semantic segmentation systems might lead to better robust transfer results, but we leave this as future work.

\input{tab/prime_with_and_without_comparison}

\input{fig/abe_data_aug_while_finetuning}

\section{Acknowledgements}
We thank the anonymous reviewers for their helpful feedback on this work. This research was in part funded by the Masason Foundation to YY.

%% file: fig/abe_degree_of_finetune.tex
\begin{figure*}
\begin{center}
\includegraphics[width=\columnwidth]{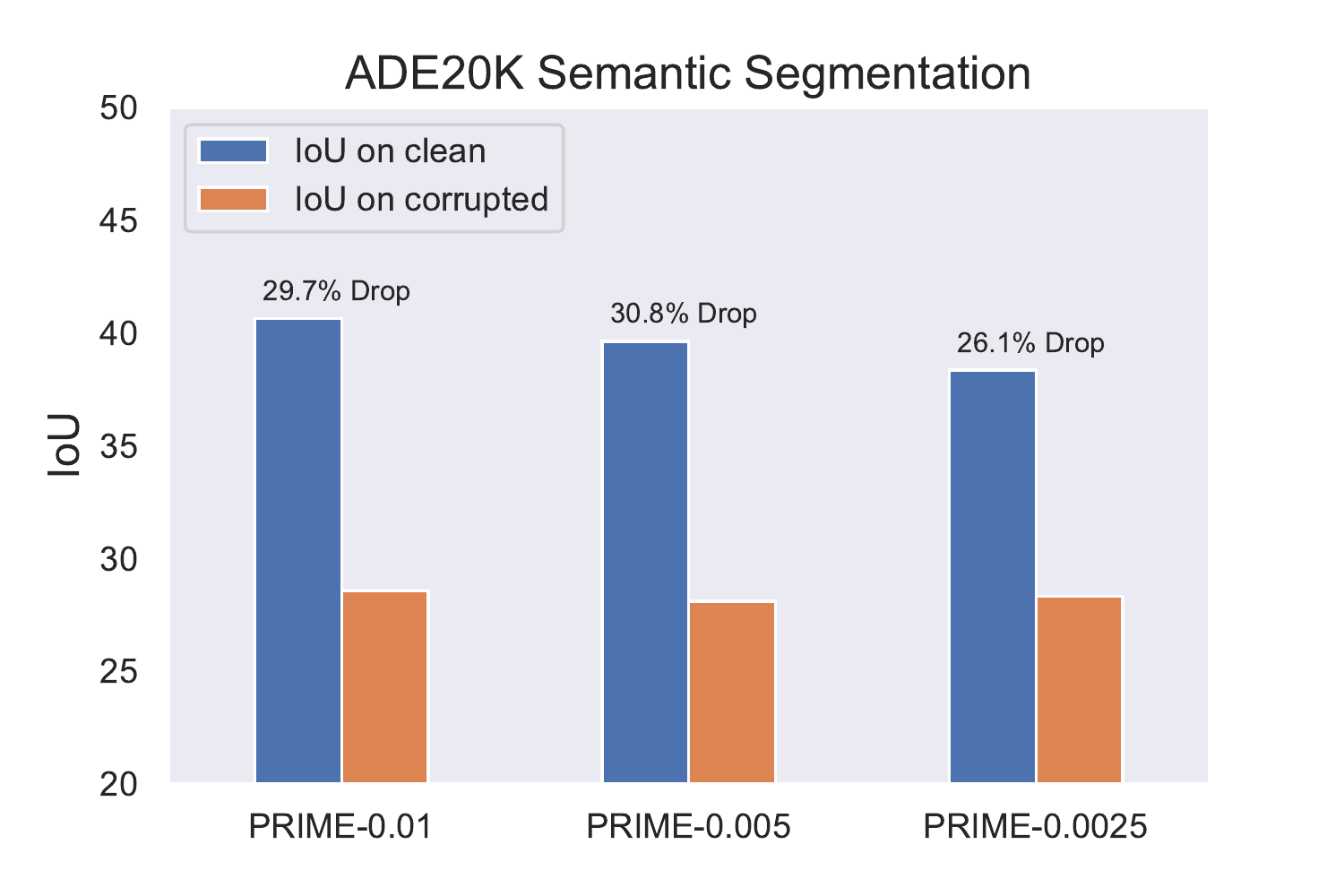}
\hfill
\includegraphics[width=\columnwidth]{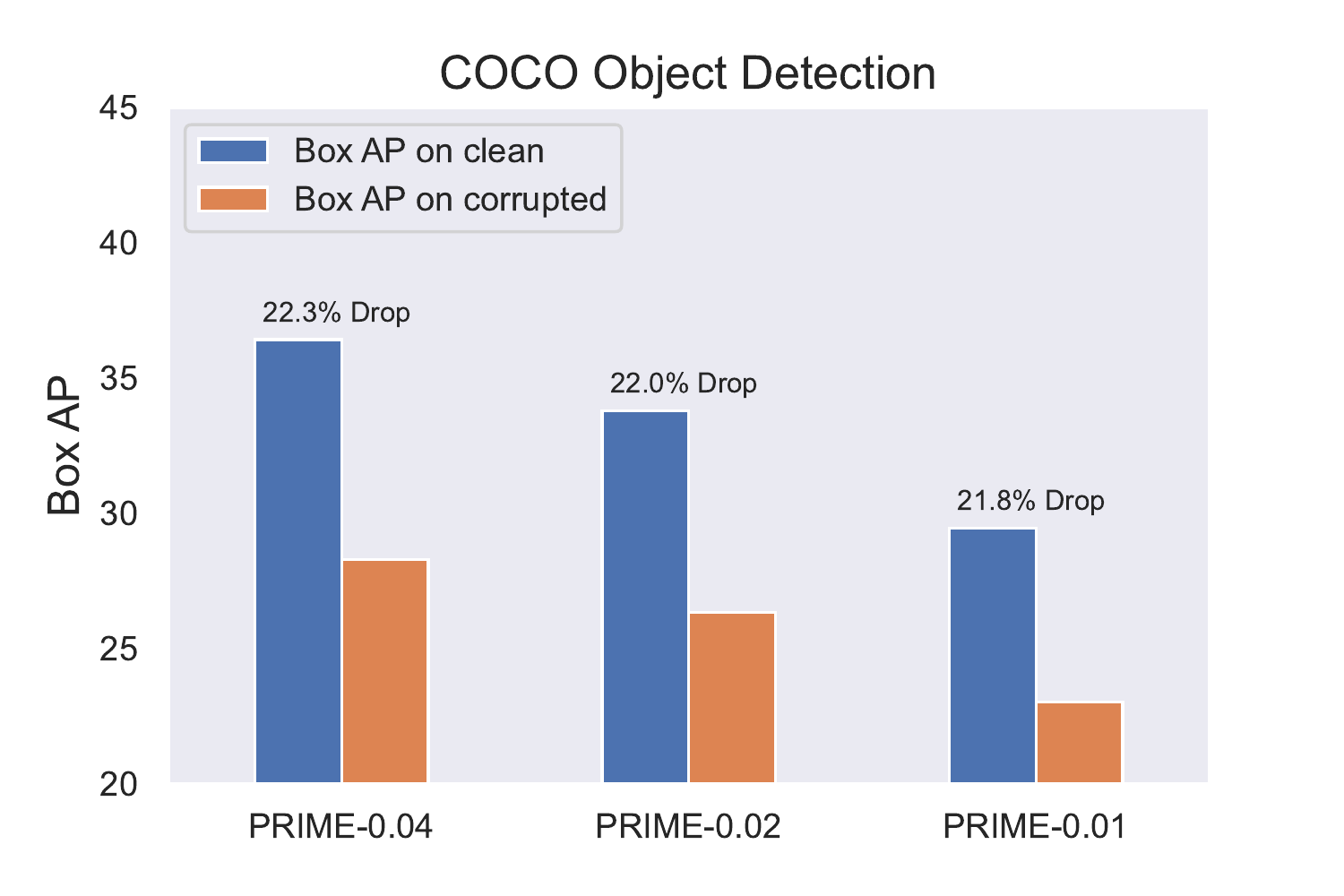}
\end{center}
\caption{
The effect of the degree of fine-tuning robust ImageNet backbones for downstream tasks. %
We use the data augmentation technique called PRIME , which corrupts images during fine-tuning.
We measure the degree of fine-tuning by the value of the learning rate we use for fine-tuning. 
We evaluate three different settings with varying learning rates.
We can see that for both ADE20K semantic segmentation and COCO object detection, the smaller the learning rate is, the lower the clean performance becomes.
However, the percentage of performance drop becomes less significant for the small learning rate, which confirms our intuition that small learning rates better retain the robustness property of the original ImageNet backbones.
}
\label{fig:abe_degree_of_finetune}
\end{figure*}

%% file: tab/prime_with_and_without_comparison.tex
\begin{table}
\centering
\resizebox{0.8\linewidth}{!}{ %
\begin{tabular}{@{}lcc@{}}
\toprule
COCO & PRIME & PR-PRIME \\
\midrule
Box AP clean  & 36.42 & 31.47  \\
Box AP corrupted & 28.30 & 24.80  \\
BoX AP Drop & 22.29 \% & 21.21 \% \\
\bottomrule
\end{tabular}
} %
\caption{
The effect of PRIME data augmentation during transfer learning from ImageNet models to COCO object detection.
Both PRIME and PR-PRIME are initialized with weights that are pretrained using PRIME on ImageNet.
For PRIME, we use a standard training for transfer learning. 
For PR-PRIME, we apply the PRIME data augmentation during transfer learning.
} %
\label{tab:prime_with_and_without_comparison_coco}
\end{table}

\begin{table}
\centering
\resizebox{0.8\linewidth}{!}{ %
\begin{tabular}{@{}lcc@{}}
\toprule
ADE20K & PRIME & PR-PRIME \\
\midrule
IoU clean  & 40.64 & 40.91  \\
IoU corrupted & 28.56 & 30.56  \\
IoU Drop & 29.73 \% & 25.31 \% \\
\bottomrule
\end{tabular}
} %
\caption{
The effect of PRIME data augmentation during transfer learning from ImageNet models to ADE20K semantic segmentation.
Both PRIME and PR-PRIME are initialized with weights that are pretrained using PRIME on ImageNet.
For PRIME, we use a standard training for transfer learning. 
For PR-PRIME, we apply the PRIME data augmentation during transfer learning.
} %
\label{tab:prime_with_and_without_comparison_ade20k}
\end{table}

%% file: fig/abe_data_aug_while_finetuning.tex
\begin{figure*}
\begin{center}
\includegraphics[width=\columnwidth]{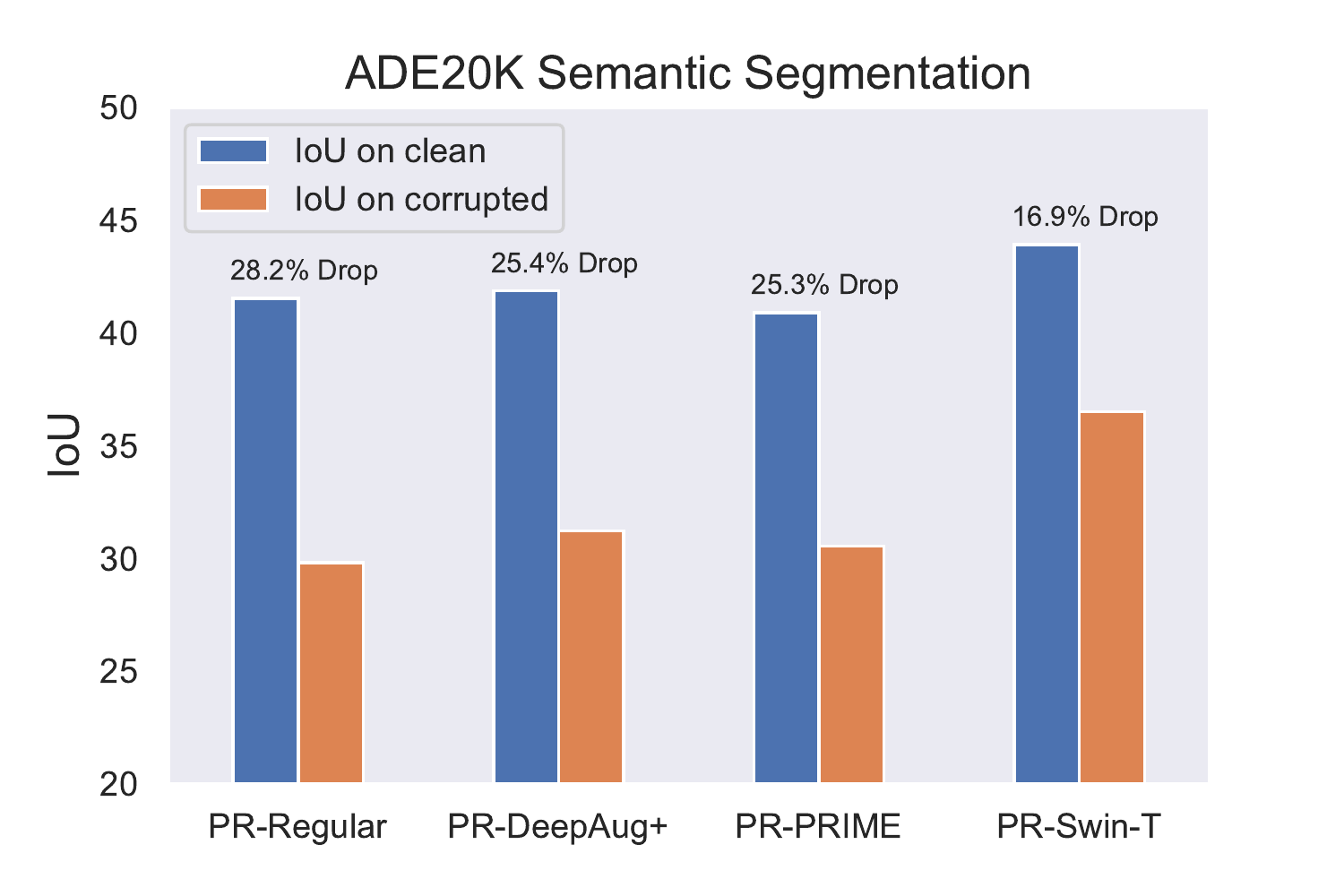}
\hfill
\includegraphics[width=\columnwidth]{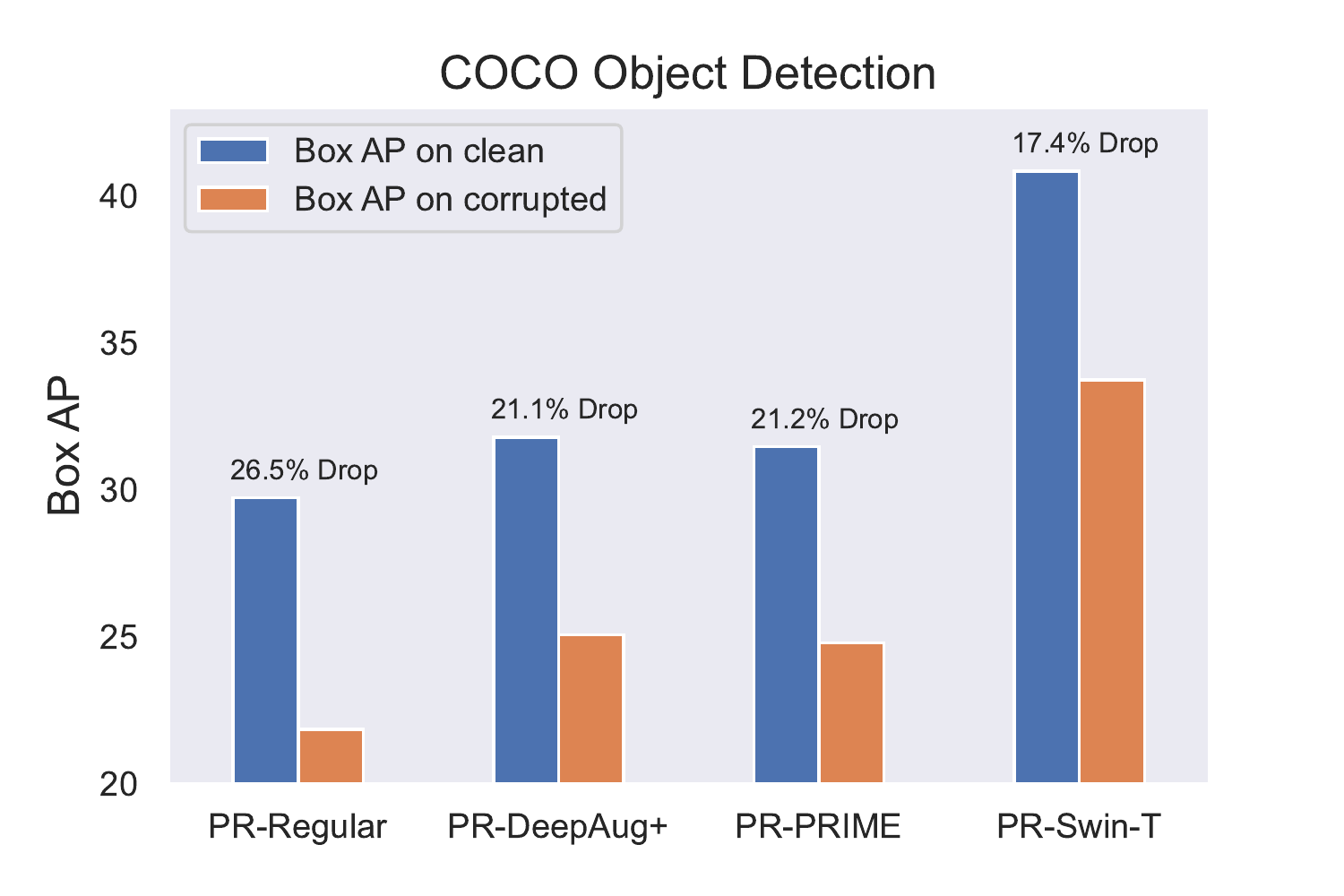}
\end{center}
\caption{
The effect of data augmentation during transfer learning from ImageNet pretrained weights to downstream tasks.
The `PR' stands for PRIME, the robustification data augmentation technique we used during fine-tuning.
We compare three ImageNet-pretrained ResNet50 models (Regular, DeepAug+, and PRIME) with Swin-T, which is pretrained on ImageNet and also use PRIME during transfer learning. During ImageNet training, DeepAug+ and PRIME use the DeepAug+AugMix and PRIME data augmentation, respectively. 
Interestingly, it seems that the PRIME fine-tuning does not affect the clean performance on the semantic segmentation task (ADE20K), although it hurts the clean performance on the COCO object detection. (See Figure \ref{fig:fullnet_compari} for comparison.)
Regarding the difference in architecture, we can see that Swin-T still performs the best in terms of both raw performance metrics on clean/corrupted data and the percentage of the performance drop.
}
\label{fig:abe_data_aug_while_finetune}
\end{figure*}